\documentclass[sn-basic]{sn-jnl}

\usepackage{tcolorbox}
\usepackage{graphicx}%
\usepackage{multirow}%
\usepackage{amsmath,amssymb,amsfonts}%
\usepackage{amsthm}%
\usepackage{mathrsfs}%
\usepackage[title]{appendix}%
\usepackage{xcolor}%
\usepackage{textcomp}%
\usepackage{booktabs}%
\usepackage{algorithm}%
\usepackage{algorithmicx}%
\usepackage{algpseudocode}%
\usepackage{listings}%
\usepackage{array}%

\theoremstyle{thmstyleone}%
\theoremstyle{thmstyletwo}%

\theoremstyle{thmstylethree}%

\begin{document}

\title[Article Title]{Halfway Escape Optimization: A Quantum-Inspired Solution for General Optimization Problems}


\author*[1]{\fnm{Jiawen} \sur{ Li}}\email{Jiawen.Li2004@student.xjtlu.edu.cn}

\author[2]{\fnm{Anwar PP} \sur{ Abdul Majeed}}\email{anwarm@sunway.edu.my}
\equalcont{These authors contributed equally to this work.}

\author[1]{\fnm{Pascal} \sur{Lefevre}}\email{Pascal.Lefevre@xjtlu.edu.cn}
\equalcont{These authors contributed equally to this work.}

\affil*[1]{\orgdiv{School of AI and Advanced Computing}, \orgname{Xi’an Jiaotong-Liverpool University}, \orgaddress{\street{Ludu Street}, \city{Suzhou}, \postcode{215123}, \state{Jiangsu}, \country{China}}}

\affil[2]{\orgdiv{School of Engineering and Technology}, \orgname{Sunway University}, \orgaddress{\street{Jalan Universiti}, \city{Bandar Sunway}, \postcode{47500}, \state{Selangor Darul Ehsan}, \country{Malaysia}}}


\abstract{This paper first proposes the Halfway Escape Optimization (HEO) algorithm, a quantum-inspired metaheuristic designed to address general optimization problems. The HEO mimics the effects between quantum such as tunneling, entanglement. After the introduction to the HEO mechansims, the study presents a comprehensive evaluation of HEO's performance against extensively-used optimization algorithms, including Particle Swarm Optimization (PSO), Genetic Algorithm (GA), Artificial Fish Swarm Algorithm (AFSA), Grey Wolf Optimizer (GWO), and Quantum behaved Particle Swarm Optimization (QPSO). The primary analysis encompasses 14 benchmark functions with dimension 30, demonstrating HEO's effectiveness and adaptability in navigating general optimization problems. The test of HEO in Pressure Vessel Design and Tubular Column Design also infers its feasibility and potential in real-time applications. Further validation of HEO in Osmancik-97 and Cammeo Rice Classification achieves a higher accuracy record.}

\keywords{Swarm intelligence, Optimization, Metaheuristics, Halfway Escape Optimization(HEO)}



\maketitle

\section*{Highlights}
\label{Highlights}
\vspace{-2ex}
\begin{tcolorbox}[colback=gray!20, colframe=gray!50]
\begin{itemize}
\item we propose a new algorithm called HEO
\item we provide a comprehensive 
benchmark of 14 Objective Functions for our HEO algorithm
\item we further evaluate the effectiveness of HEO in 3 classical engineering problems
\item we optimize Logistic Regression model with HEO, achieving a higher record in Osmancik-97 and Cammeo Rice Classification
\end{itemize}
\end{tcolorbox}

\section{Introduction}
\label{introduction}
The main motivation behind the development of the Halfway Escape Optimization (HEO) algorithm stems from the limitations of existing optimization methods in terms of efficiency and adaptability to various industrial single-objective optimization problems. 

HEO is a quantum-inspired metaheuristic designed to tackle the distinction between different search spaces in industrial scenarios with a focus on achieving fast convergence rates and reducing the time cost in searching. Inspired by the behavior of quantum particles and the concept of halfway escape, the algorithm offers a versatile and adaptive approach to exploration and exploitation in challenging optimization domains.

This paper provides a comprehensive analysis to the HEO algorithm, focusing on its core principles, adaptability, and robustness in navigating diverse optimization landscapes. The unique energy-driven behavior, vibration strategies, and exploratory mechanisms of HEO are examined in the context of solving a range of simple as well as complex benchmark functions. The evaluations demonstrate HEO's effectiveness in achieving convergence, balancing exploration and exploitation, and discovering high-quality solutions across a variety of challenging optimization problems.

Comparative studies with other swarm optimization algorithms highlight HEO's adaptability and robustness in addressing the complexities of multimodal and high-dimensional search spaces, showcasing its fast convergence speed. The results of this study underscore the importance of the HEO algorithm as a promising approach to address various optimization challenges, with implications for a wide range of practical applications, including structural design\cite{structure}\cite{structure2} and model optimization\cite{model}.

The detailed analysis presented demonstrates the potential of the HEO algorithm as a powerful and adaptive optimization method.  Overall, the paper first proposes HEO method and primary testing its effectiveness in 14 single objective functions as well as 3 engineering problems including Pressure Vessel Design, Tubular Column Design, Osmancik-97 and Cammeo Rice Classification.

\section{Related Work}
\label{Related Work}
In the realm of optimization, metaheuristic algorithms play a key role in the addressing of challenging optimization problems. These algorithms draw inspiration from natural phenomena and collective behaviors to navigate high-dimensional and multimodal search spaces, offering adaptive and robust solutions and widely using in real-world optimization problems such as scheduling or logistics problems \cite{engineerProblem}.

In 1975, \citeauthor{hollandref} created a new metaheuristic algorithm called Genetic Algorithm (GA), a prominent evolutionary algorithm, that simulates natural selection and genetic recombination to evolve a population of candidate solutions. GA's adaptability has made it a go-to choice for a wide range of optimization problems. However, its exploration of high-dimensional and multimodal search spaces may be limited, potentially leading to suboptimal solutions.

Particle Swarm Optimization (PSO) has gained widespread recognition as a bio-inspired metaheuristic algorithm rooted in the collective behavior of bird flocks. PSO was first proposed by \citeauthor{eberhartref} in 1995. PSO iteratively updates the positions and velocities of the particles based on their best individual and global solutions, enabling effective exploration and exploitation of the search space\citep{eberhartref}. While PSO has exhibited robustness in addressing various optimization problems, its potential for premature convergence and challenges in handling various multimodal landscapes have been noted\citep{vazquezref}.

Ant Colony Optimization (ACO) is an algorithm inspired by the foraging behavior of ants. It was initially proposed by Marco Dorigo in his doctoral thesis in 1992 and has since gained significant attention in the field of optimization\citep{dorigoref}. ACO simulates the behavior of ants in finding the shortest path between their nest and food source by depositing pheromone trails. These trails act as a form of communication, allowing ants to navigate and explore the search space effectively. ACO has been successfully applied to a wide range of optimization problems, including the well-known Traveling Salesman Problem (TSP)\cite{path} and Vehicle Routing Problem (VRP)\citep{dorigoref}. Its ability to find near-optimal solutions and its adaptability to dynamic environments make ACO a valuable algorithm for solving complex optimization problems.

Differential Evolution (DE) is another population-based stochastic optimization algorithm that was introduced by \cite{stornref}. DE operates by iteratively evolving a population of candidate solutions through a combination of mutation, crossover, and selection operations. It is inspired by the process of natural evolution and survival of the fittest. DE utilizes the differential operators to explore and exploit the search space efficiently. It has been widely applied to various optimization problems and has shown promising performance. The simplicity, effectiveness, and robustness of DE make it a popular choice for solving real-world optimization problems.

Artificial Fish Swarm Algorithm (AFSA) emulates the foraging behavior of fish in a swarm to explore the search space and update candidate solutions\citep{fishswarmref}. AFSA has demonstrated effectiveness in continuous and combinatorial optimization tasks by adopting swarm and follow behaviors, yet its ability to search may present challenges due to its reliance on fish movement and interaction dynamics.

The Firefly Algorithm (FA) is a nature-inspired optimization algorithm proposed by \citeauthor{fireflyref} in 2009. Drawing inspiration from the flashing patterns of fireflies, FA simulates the social behavior of fireflies to solve optimization problems. Each firefly's light emission represents a potential solution, and the attractiveness between fireflies is determined by the brightness of their light and their distance from each other. Fireflies seek to improve their positions by moving towards brighter fireflies in the search space. FA has demonstrated effectiveness in solving a variety of optimization problems, including continuous, discrete, and multi-modal functions. 

In 2009, Quantum behaved Particle Swarm Optimization (QPSO),or Q-PSO,was proposed to solving multi-objective problem \cite{quantum}. QPSO is a variant of Particle Swarm Optimization (PSO) that incorporates concepts from quantum mechanics, such as quantum probabilities and superposition, to enhance its search capabilities. It aims to find optimal solutions for multi-objective problems by leveraging the principles of quantum mechanics and the collaborative behavior of a swarm of particles.

The Grey Wolf Optimizer (GWO) draws inspiration from the hierarchical structure and hunting behavior of grey wolves in nature, which suggested by \citeauthor{mirjaliliref} in 2014. Through the alpha, beta, and delta wolf concept, GWO dynamically updates solution positions, showcasing promise in addressing complex optimization problems. Nonetheless, the algorithm's sensitivity to parameter settings and the need for domain-specific fine-tuning have been identified as areas for consideration \citeauthor{mirjaliliref2}.

The Salp Swarm Algorithm (SSA) draws inspiration from the collective behavior of salps in the ocean\citep{mirjaliliref2}. This bio-inspired optimization technique emulates the swarming and foraging dynamics of salps to navigate complex search spaces and seek optimal solutions. By simulating the intricate swimming and feeding patterns of salps, SSA exhibits a remarkable capacity for exploration and exploitation, making it proficient in addressing a wide range of optimization problems, including both unconstrained and constrained scenarios. 

There still have other meta-heuristic algorithms, those algorithms could classified as two category: bio-inspired and the nature-inspired, nature-inspired means the optimization algorithms are inspired from physical or chemical laws in nature. Bio-inspired optimizations algorithms including Whale Optimization Algorithm(WOA)\cite{Whale}, Artificial Bee Colony(ABC)\cite{Bee}, Bat Algorithm(BA)\cite{Bat}, Cuckoo Search(CS)\cite{Cuckoo},Dragonfly Algorithm (DA)\cite{Dragonfly}, Flower Pollination Algorithm(FPA)\cite{Flower}, Cheetah Optimizer (CO)\cite{Cheetah}, Genetic Programming (GP)\cite{GP}, Fox Optimizer(FOX)\cite{Fox}, GOOSE algorithm\cite{GOOSE}. 

Nature-inspired optimization algorithms contains Snow Ablation Optimizer (SAO)\cite{Snow},Light Spectrum Optimization (LSO)\cite{Light}, Nuclear Reaction Optimization (NRO)\cite{Nuclear},Water Cycle Algorithm (WCA)\cite{Water},Chernobyl Disaster Optimizer (CDO)\cite{Chernobyl}. Those bio-inspired as well as nature-inspired algorithms both have contributed significantly to the field of optimization, each with its unique strengths.

\section{Proposed Methodology}
\label{Proposed Methodology}
In recent years, swarm optimization algorithms have been widely applied in various fields, such as machine learning, engineering design. These algorithms have achieved remarkable success in solving optimization problems and have shown great potential in handling complex tasks. 

However, these algorithms often face difficulties when faced with complex landscapes. One example of a group optimization algorithm facing a rugged landscape is the Particle Swarm Optimization (PSO) algorithm. When facing rugged landscapes with many local optima, PSO often gets trapped in these local optima and fails to find the global optimum\cite{vazquezref}. 

The difficulty in finding global optima in rugged landscapes can be attributed to several reasons. First, the rugged landscape creates many false peaks, making it challenging for the optimization algorithm to distinguish between the local optima and the global optimum. Second, the steep gradients in rugged landscapes lead to rapid convergence to local optima, which hinders the exploration of the search space \cite{gradient}. Third, the presence of many local optima increases the probability of particles getting trapped in these optima. Especially in high-dimensional situations, those hollow spaces would further reduce the chances of finding the global optimum \cite{cai2020solution}.

To solve those issues, the paper designed HEO (Halfway Escape Optimization) based on a few behaviors from quantum. HEO algorithm includes four different mechanisms for exploitation(\ref{lab:clip} Center Clipping) as well as exploration (\ref{lab:pos} Position Update, \ref{lab:vib} Vibration, \ref{lab:skip} Random Skip). 

 Yet QPSO and HEO are both quantum-inspired algorithms with the same mechanism in computing global weight for updating position(seen in equation \eqref{eq:6}), they still have significant differences. In contrast to the QPSO, QPSO only allows for a single state change, HEO is more inclined to escape when convergence is achieved, resulting in the inclusion of velocity during the escape phase. Moreover, HEO incorporates random searches not only within the range of updates but also for the entire swarm. Additionally, utilizing a center clipping strategy in HEO contributes to its superior convergence rate.

The Position Update mechanism, which forms the basis of all swarm optimization algorithms, which are the moving way of each individual entity in the swarm, is designed differently in HEO. Instead of focusing on a single peak, the quantum in HEO has two states that determine whether it should approximate the optimum. This unique way of updating positions allows the swarm in HEO to escape from false peaks, thus addressing the first difficulty mentioned earlier. Like the quantum could shuttle through the potential energy barrier \cite{tunneling}, the movement of the quantum in HEO is not influenced by gradients like in the PSO algorithm, making it more stable and less likely to be hindered by complex landscapes. Furthermore, in order to navigate complex hollow landscapes, HEO incorporates mechanisms such as increasing the escape factor, Vibration, and Random Skip to accelerate the quantum's ability to overcome obstacles. Lastly, for exploring complex situations, the Center Clipping mechanism in HEO ensures that the swarm is relatively concentrated around the already discovered optima, thereby efficiently searching for solutions in a convex environment.

\subsection{HEO Overview}
\begin{figure}[H]
    \centering
    \includegraphics[width=0.9\linewidth]{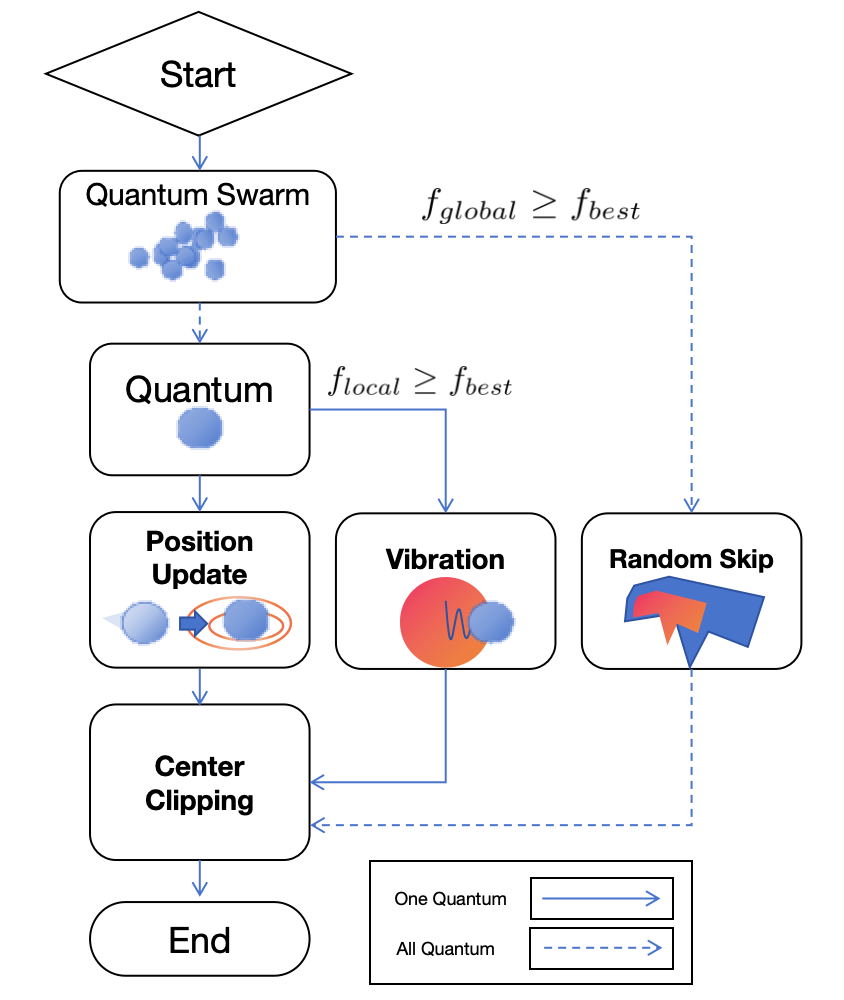}
    \caption{Simplified Flowchart of One Iteration in HEO}
    \label{fig:flow}
\end{figure}

As Fig.\ref{fig:flow} shown, in the Halfway Escape Optimization (HEO) algorithm, the swarm is structured as a collection of quantum entities. Each quantum entity iteratively adjusts its position and employs the Center Clipping strategy to satisfy the defined constraints while enhancing the efficiency of the search process. Should the swarm fail to identify a superior global optimum, a random skip is initiated, allowing for a departure from the current trajectory. Concurrently, individual quantum entities engage in a vibration mechanism when local optima are not improved. This action aims to explore beyond the plateaus or local convex hulls, facilitating a more comprehensive exploration of the search space. Details of HEO can be seen in \textbf{Algorithm 1}.

\begin{algorithm}[H]
\caption{Halfway Escape Optimization (HEO)}
\begin{algorithmic}
\Procedure{HEO}{$k, i_{max}, a_{\text{max}},c_{\text{max}}, f, p, \text{bound}, R$}
    \State $Q \gets$ list of $k$ quantums with positions,levels and local optimas
    \State $f_{\text{best}} \gets \min(\{f(q.\text{x}) \mid q \in Q\})$, $c_{i} \gets 0$
    \State $q_{\text{best}} \gets$ quantum in $Q$ with fitness $f_{\text{best}}$
    \For{$i \gets 1$ to $i_{max}$}
        \ForAll{$q$ in $Q$}
          \State update position of $q$
            \State $f_{q} \gets f(q.\text{x})$
            \If{$f_{q} < f_{\text{best}}$}
                \State Update $f_{\text{best}},q_{\text{best}}$ 
                \State $c_{i} \gets int(c_{i} / 2$)
            \ElsIf{$f_{q} < f(q.\text{local\_best}.\text{x})$}
                \State Update $q.local\_best$
                \State $q.a_{i} \gets int(q.a_{i} / 2$)
            \Else
                \State $n \gets N(0,\sigma_{q.x},p)$
                \State $q.x$+=$\frac{n}{1+e^{a_{i}}}$
            \EndIf
            \State center clipping $q.\text{x}$ in $S_{result}$
        \EndFor
        \If{$q.a_{k}*r_{4}<\frac{a_{max}-1}{2}$}
                \State $q.a_{k}$++
        \EndIf
        \If{$c_{i}>c_{max}$}
            \State every quantum $q$ random skip
            \State $c_{i} \gets 0$
        \EndIf
        \State $c_{i}$++
    \EndFor
    \State \Return $f_{\text{best}}$, $q_{\text{best}}$
\EndProcedure
\end{algorithmic}
\end{algorithm}

\subsection{Position Update}
\label{lab:pos}

\begin{figure}[H]
    \centering
    \includegraphics[width=0.8\linewidth]{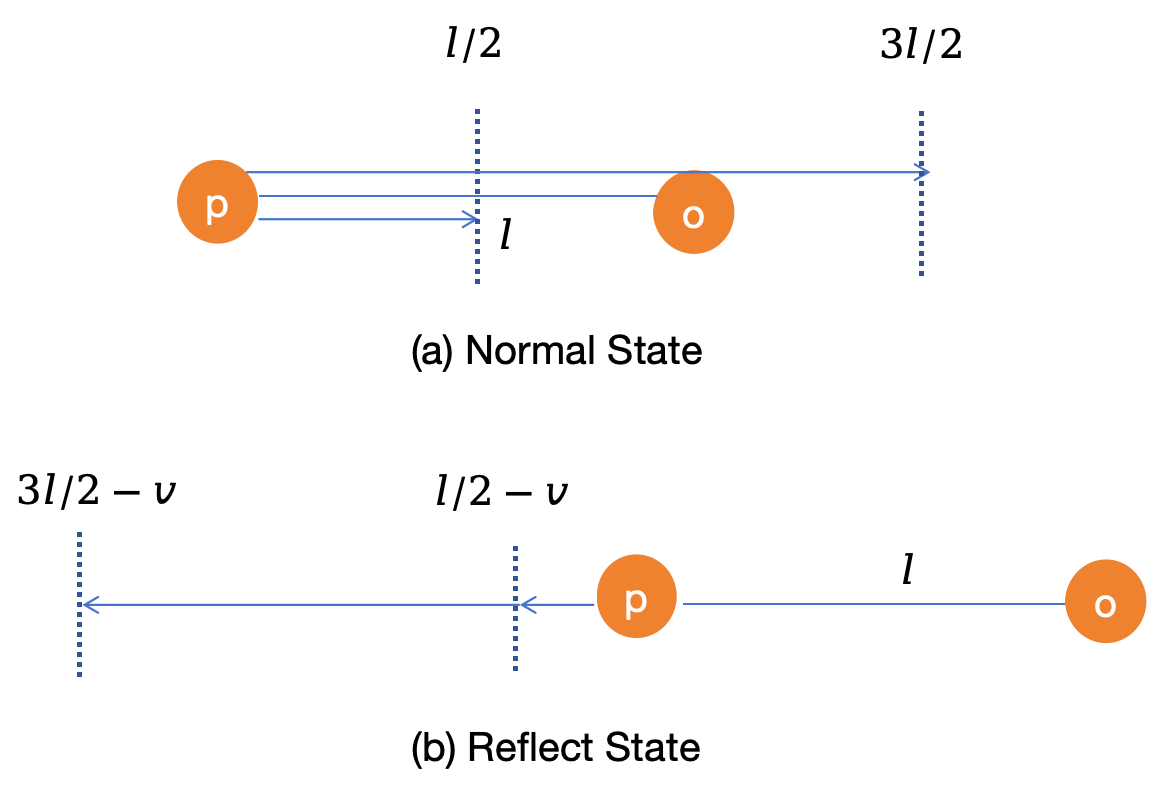}
    \caption{Position Update in HEO}
    \label{fig:pos}
\end{figure}

The mechanism Position Update of HEO, as it is named, is the special way that quantum updates its position for search, which is the fundamentals of all optimization algorithms. Similar to PSO, in HEO, the updates of quantums' position consider both the local optimum and global optimum. The core concept of the HEO is "escape", the quantum in HEO would move towards the opposite direction once the group gets stuck in the local optima, this trait allows HEO to escape from it. First of all, as Fig.\ref{fig:pos} shows, the HEO is not fully gradient dependent like PSO, HEO only uses fitness or cost for updating the velocity of escape as well as updating the bound of the quantum, which makes HEO find relatively good solutions on the sheer and other narrow landscapes. 

\begin{equation}
\label{eq:1}
    \bold{x_{i+1}}=\bold{x_{i}}+\bold{v_{g}}+\bold{v_{l}} ,\quad i=1,2,3...N\tag{1}
\end{equation}

\begin{equation}
\label{eq:2}
    \bold{v_{g}}=(\bold{x_g}-\bold{x_{i}}(c_{i}+1)\times r_{1})\times r_{2}\times r_{3} \tag{2}
\end{equation}
\begin{equation}
\label{eq:3}
    \bold{v_{l}}=(\bold{x_l}-\bold{x_{i}}(c_{i}+1)\times r_{1})\times r_{2}(1-r_{3}) \tag{3}
\end{equation}

In the normal state(a), quantum does not have any velocity. When the escape factor $c_{i}$ is equal to zero the behavior is equivalent to finding the solution surrounding the halfway of optima in equation \eqref{eq:1}.To be precise, the HEO algorithm would get into reflect state(b) as Fig.\ref{fig:pos} shows, to find other possible solution spaces. Unlike directly searching randomly, the reflecting state of the HEO's quantum utilizes the information(which is $c_{i}$ here) from local optima, allowing it to escape from it faster. 

\begin{equation}
\label{eq:4}
r_{1} \sim U(1-R, 1+R) \tag{4}
\end{equation}

\begin{equation}
\label{eq:5}
r_{2} \sim U(0.5, 1.5)  \tag{5}
\end{equation}

\begin{equation}
\label{eq:6}
r_{3} \sim U(0, 1)  \tag{6}
\end{equation}

\begin{equation}
\label{eq:7}
c_{i+1}=
\left\{
\begin{aligned}
0 \quad \quad \quad \quad \quad \quad
(\text{when }i=0)\\
c_i+1 \quad (\text{when } f_{global}>f_{best})\\
c_i\mid 2 \quad \quad \quad \quad 
\quad \quad
\text{else} \quad \quad \quad
\end{aligned}
\tag{7}
\right.
\end{equation}

As for the random factors $r_{1}$ and $r_{2}$ shown in equation \eqref{eq:4} and equation \eqref{eq:5}, $r_{1}$ is used to control the magnitude of an escape step, making the HEO escape behaviors more efficient than the simple grid search.$r_{2}$ is added to further randomize the process. Both of those two random variables have exception 1 to ensure the feasibility. The random factor $r_{3}$ controls the global weight of this update process, creating a fan area to search for the solution.

\subsection{Vibration}
\label{lab:vib}
Another behavior of HEO is common in other algorithms, including AFSA, when the search is meaningless to make a random step for an individual search. Yet those algorithms adopt the strategy, they ignore the influence of size of steps, and those search approaches highly rely on the hyper-parameters tunning. HEO tries to use the standard deviation of positions to judge the magnitude of one random step, creating a relatively stable performance compared to the others, this random search is called Vibration(seen in equation\eqref{eq:13}. 

\begin{equation}
\label{eq:13}
    \bold{x_{i+1}}=\bold{x_{i}}+\frac{\bold{n}}{1+e^{a_{i}}} \tag{8}
\end{equation}

\begin{figure}[H]
    \centering
    \includegraphics[width=0.8\linewidth]{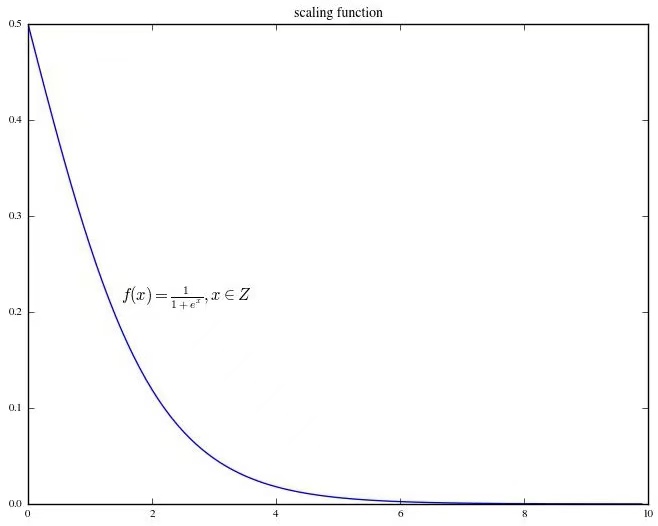}
    \caption{Scaling Function}
    \label{fig:scale}
\end{figure}

\begin{equation}
\label{eq:14}
a_{i+1}=
\left\{
\begin{aligned}
a_i+1 \quad
(\text{when } a_i*r_{4}<a_{max}) \\
a_i\mid 2 \quad (\text{when } f_{local}<f_{best})
\end{aligned}
\tag{9}
\right.
\end{equation}

\begin{equation}
\label{eq:15}
r_{4} \sim U(0, 1) \tag{10}
\end{equation}

\begin{equation}
\label{eq:16}
\bold{n} \sim N(0,\sigma_{x_{i}})  \tag{11}
\end{equation}

Just like the property of quantum in physics, with higher energy level $a_{i}$, the quantum would be harder get influenced by others\citep{sommref}, which means smaller oscillations(seen in equation \eqref{eq:13} and Fig.\ref{fig:scale}), maintain an extent of diversity in the population. To maintain the convergence of the whole group, the increment of energy level $a_{i}$ only works in a few quantum. As for amplitude of vibration, other scaling approaches to adapting the space of high dimensions with gamma functions, are computed much slower than just simply calculating the standard deviation of the position vectors as shown in equation\eqref{eq:16}. The reason for using Normal distribution due its universal applicability. According to the central limit theorem, with the increasing of the dimension, the size of the vibration is more approximate to the size of the step in the real situation. Since the HEO does not include periodical functions in this behavior, it cannot be defined as oscillations, that's the reason for its name. \\

\subsection{Center Clipping}
\label{lab:clip}

Just like quantum entanglement causes quantum attraction, so the distance of quantum escape is finite, the quantum group in HEO would restrict the individuals, ensuring them in a certain area to avoid over-escape by the behavior.

\begin{equation}
\label{eq:17}
S_{result} = S_{bound} \cap S_{group} \tag{12}
\end{equation}

\begin{equation}
\label{eq:18}
b_{g} = \|\bold{x_{i}}-\bold{x_{g}}\|_2 \times r_{5} \tag{13}
\end{equation}

\begin{equation}
\label{eq:19}
r_{5} \sim U(0, 2) \tag{14}
\end{equation}

\begin{figure}[h]
    \centering
    \includegraphics[width=0.95\linewidth]{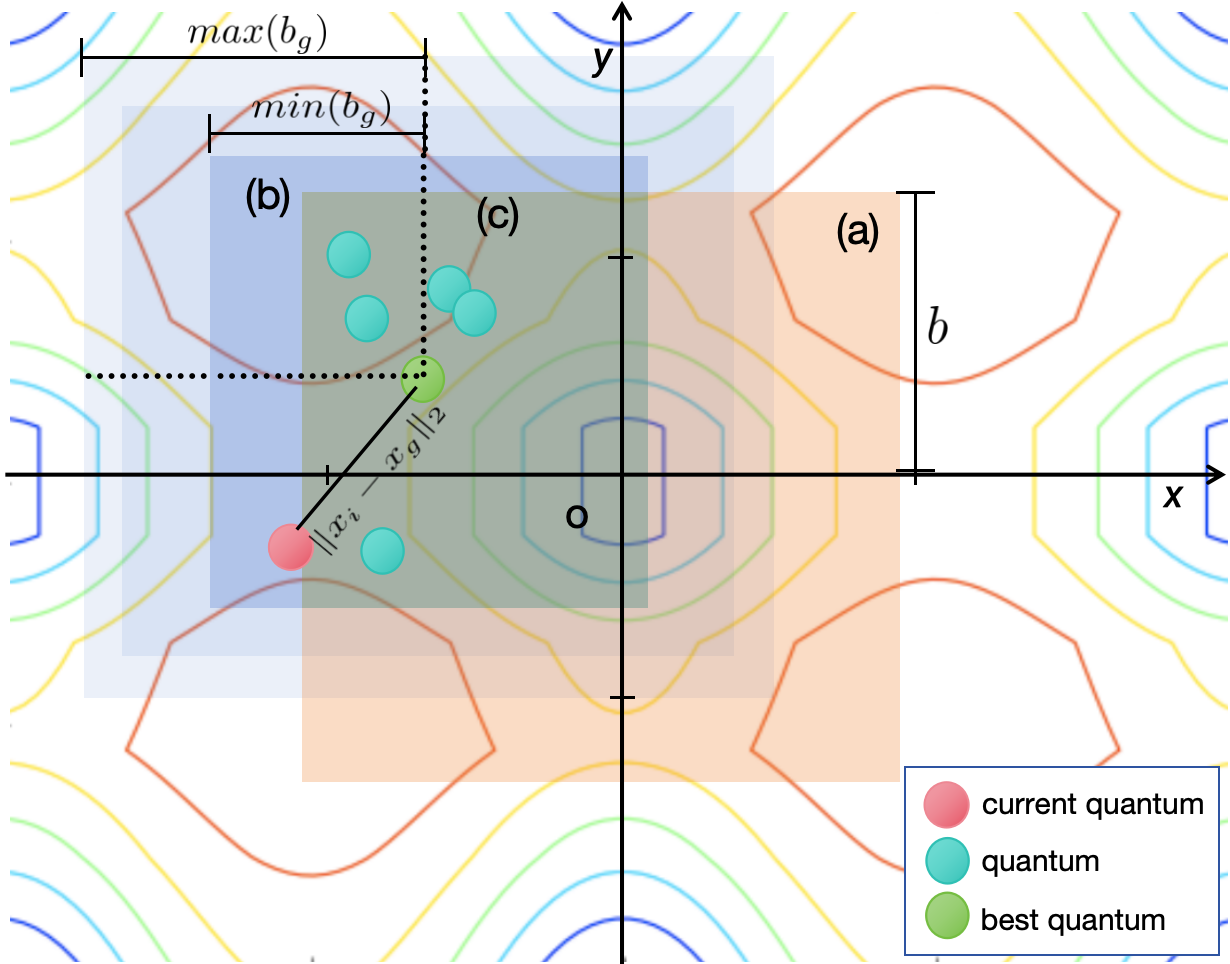}
    \caption{Center Clipping in Ackley Function}
    \label{fig:clip}
\end{figure}

 As Fig.\ref{fig:clip} illustrated, the quantum not only would be restricted by the search area or constraints $S_{bound}$(the area with (a)) with half of the side length $b$, but the group also would be restricted by $S_{group}$. $S_{group}$(the area of (b) in Fig.3)is the search space of the group that HEO allows to search with half of the side length $b_{g}$, the search space in HEO are all in the shape of a square, making it computing faster than calculating sphere. Moreover, as equation \eqref{eq:18} shows, this bound of group $S_{group}$ depends on the differences between the current quantum's position and the best one that HEO found. This center clipping mechanism uses a random factor $r_{5}$, which makes this bound soft edge not converge too early and maintains the diversity of the group. The final result of the group gets into the range of $S_{result}$(area (c)), which is the overlap area of the $S_{group}$(area (b)) and $S_{bound}$(area (a)).
 
\subsection{Random Skip}
\label{lab:skip}
The HEO is a swarm algorithm extremely focused on exploration. Except for the vibration behavior that each quantum has, the group itself would be randomly skipped to search the solution with the range of the bound. In equation \eqref{eq:19}, once the count $c_{i}$ reached $c_{max}$ and then manually defined, the whole group would make this random skip for spanning a totally new space. The reason to use this skipping mechanism is the higher randomness compared to using any periodical function to search in HEO.

\begin{equation}
\label{eq:20}
\bold{x_{i}}= \frac{\bold{x_{i}}+\bold{\Vec{r}}}{2}\tag{15}
\end{equation}

\begin{equation}
\label{eq:21}
\Vec{r}_{j} \in \bold{\Vec{r}} \tag{16}
\end{equation}

\begin{equation}
\label{eq:22}
\Vec{r}_{j} \sim U(0, \bold{b}) \tag{17}
\end{equation}

The random skip behavior in HEO allows the entire group of quantum to explore a completely new and random search space, the blue area represents the search space, and the orange part is for the constraint. This is achieved by randomly shifting the position of each quantum by a random vector $\Vec{r}$ in Equation \eqref{eq:20}. The components of the random vector $\Vec{r}_{j}$ are uniformly sampled from the interval $[0, b]$, as shown in equation \eqref{eq:22}. There are numerous $b$ for $\Vec{r}$, hence the shape of the search space also changed with it, also the space restricts the constraints. This random skip mechanism enables HEO to explore different regions of the search space and potentially discover new and better solutions.

\section{Experimental Results}
\label{Experimental Results}

\subsection{Experiment Design}
The paper selects 14 benchmark functions with a relatively laerge bound within -100 to 100(the reason to set a same bound is to test the adaptability of each algorithms) from famous objective functions that have extensional forms in high dimensions, including 7 unimodal functions in TABLE \ref{tab:uni} and 7 multimodal functions in TABLE \ref{tab:multi}. The experiment would test the performance of classical metaheuristic algorithms including PSO, AFSA, GWO, and GA with the HEO algorithm (check the standard HEO in \textbf{Algorithm 1}) in dimension $p=30$, and searching within 1000 iterations. Each swarm of the testing algorithms have 100 entities, which is common size of the swarm for testing as a rule of thumb. In addition, the experiment is validated 30 times to remove the potential impact caused by randomness.

\begin{table}[h]
\caption{Test Unimodal Functions}%
\begin{tabular}{@{}lllll@{}}
\toprule
\textbf{Name} & \textbf{Function} & \textbf{range} & \textbf{$f_{min}$} \\
\midrule
Sphere & $F_{1}(x) = \sum_{i=1}^{n} x_i^2$&[-100,100]&0 \\
\hline
Step & $F_{2}(x) =  \sum_{i=1}^{n} (x_i + 0.5)^2$&[-100,100]&0 \\
\hline
Schwefel 2.21 & $F_{3}(x) =  \max_i |x_i|$&[-100,100]&0 \\
\hline
Schwefel 2.22 & $F_{4}(x) =  \sum_{i=1}^{n} |x_i| + \prod_{i=1}^{n} |x_i|$&[-100,100]&0 \\
\hline
Rosenbrock & $F_{5}(x) =   \sum_{i=1}^{n-1} \left[ (x_{i+1} - x_i^2)^2 + (x_i - 1)^2 \right]$&[-100,100]&0 \\
\hline
BentCigar & $F_{6}(x) = x_1^2 + 10^{6} \sum_{i=2}^{n} x_i^2$&[-100,100]&0 \\
\hline
Sumsquares2 & $F_{7}(x) =  \sum_{i=1}^{n} i \cdot x_i^2$&[-100,100]&0 \\
\botrule
\end{tabular}
\label{tab:uni}
\end{table}

\begin{table}[h]
\caption{Test Multimodal Functions}%
\begin{tabular}{@{}l>{\raggedright\arraybackslash}p{7cm}lll@{}}
\toprule
\textbf{Name} & \textbf{Function} & \textbf{range} & \textbf{$f_{min}$} \\
\midrule
Alpine & $F_{8}(x) = \sum_{i=1}^{n} \left| x_i \cdot \sin(x_i) + 0.1x_i \right|$&[-100,100]&0 \\
\hline
Griewank & $F_{9}(x) =   \frac{1}{4000} \sum_{i=1}^{n} x_i^2 - \prod_{i=1}^{n} \cos\left(\frac{x_i}{\sqrt{i}}\right) + 1$&[-100,100]&0 \\
\hline
Rastrigin & $F_{10}(x) =   \sum_{i=1}^{n} \left( x_i^2 - 10 \cos(2\pi x_i) + 10 \right)$&[-100,100]&0 \\
\hline
Ackley & $F_{11}(x) =  20 + e - 20 \exp\left(-0.2 \sqrt{\frac{1}{n} \sum_{i=1}^{n} x_i^2}\right) - \exp\left(\frac{1}{n} \sum_{i=1}^{n} \cos(2\pi x_i)\right)$&[-100,100]&0 \\
\hline
L$\acute{e}$vy & $F_{12}(x) = \sin^2(\pi w_1) + \left(\sum_{i=1}^{n-1}  (w_i - 1)^2 \left(1 + 10 \sin^2(\pi w_i + 1)\right)\right)+ (w_n - 1)^2 \left(1 + \sin^2(2\pi w_n)\right), w_i = 1 + \frac{x_i - 1}{4}$
&[-100,100]&0 \\
\hline
Salomon &$F_{13}(x) = 1-\cos(2 \pi \sqrt{\sum_{i=1}^{n} x_{i}^{2} })+0.1\sqrt{\sum_{i=1}^{n} x_{i}^{2} }$&[-100,100]&0 \\
\hline
Schaffer &$F_{14}(x) = 0.5 + \frac{1}{D-1}\sum_{i=1}^{D-1}(\sin(\sqrt{x_i^2 + x_{i+1}^2})^2 - 0.5)
 $&[-100,100]&0 \\
\botrule
\end{tabular}
\label{tab:multi}
\end{table}

\subsection{Trend Analysis}

\begin{figure}[H]
    \centering
    \includegraphics[width=1\linewidth]
    {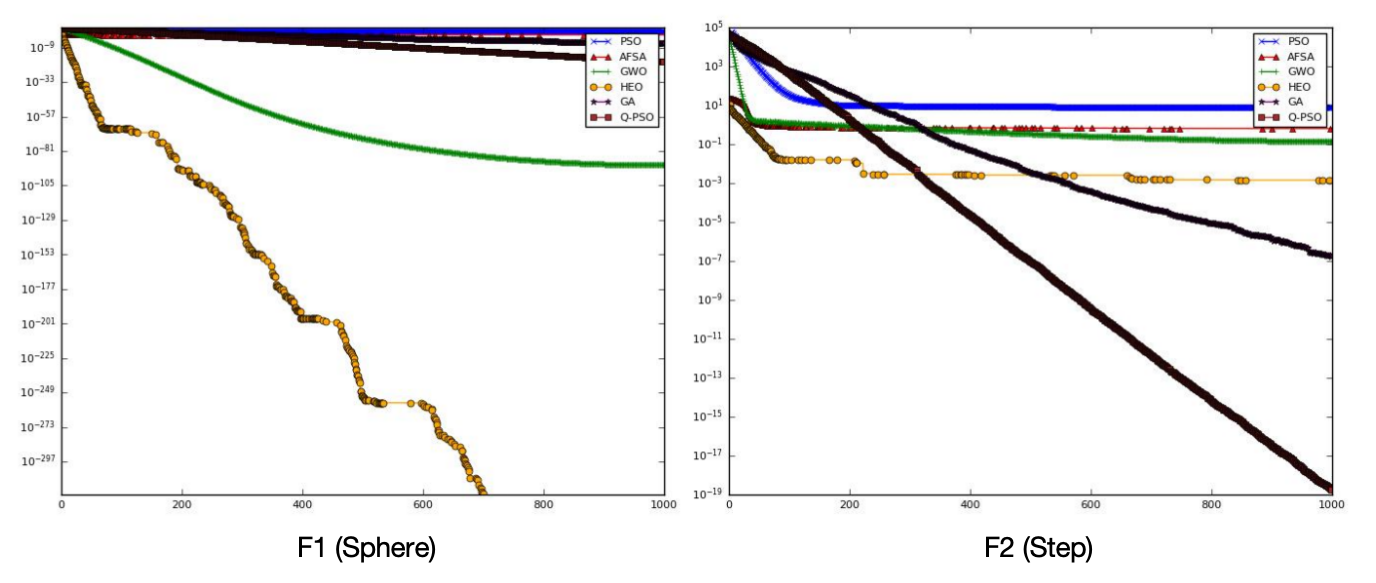}
    \caption{The cost variation in $F_{1}$ and $F_{2}$}
    \label{fig:F1-F2}
\end{figure}

In Fig.\ref{fig:F1-F2}, the HEO achieves 0 costs and converges faster than other algorithms in $F_{1}$. The GWO gets the second low-cost. As Table \ref{tab:4.3} indicates, HEO gets the 0 cost and GWO gets the 1.651e-91 cost, which could also be considered to be almost the global optimum. 

The Fig.\ref{fig:F1-F2}  demonstrates the QPSO gets the smallest cost among the others, and the GA and HEO are next in $F_{2}$. As for the convergence, except QPSO, the HEO has get relatively low cost in the beginning to about the 600 iterations. After that, the HEO's performance is exceeded by the GA. Also, the slope of the GA plot infers the acceptable convergence in further searching, it infers the single constraint conditions might make the HEO hard to converge due to the frequent skip and vibration.

\begin{figure}[H]
    \centering
    \includegraphics[width=1\linewidth]
    {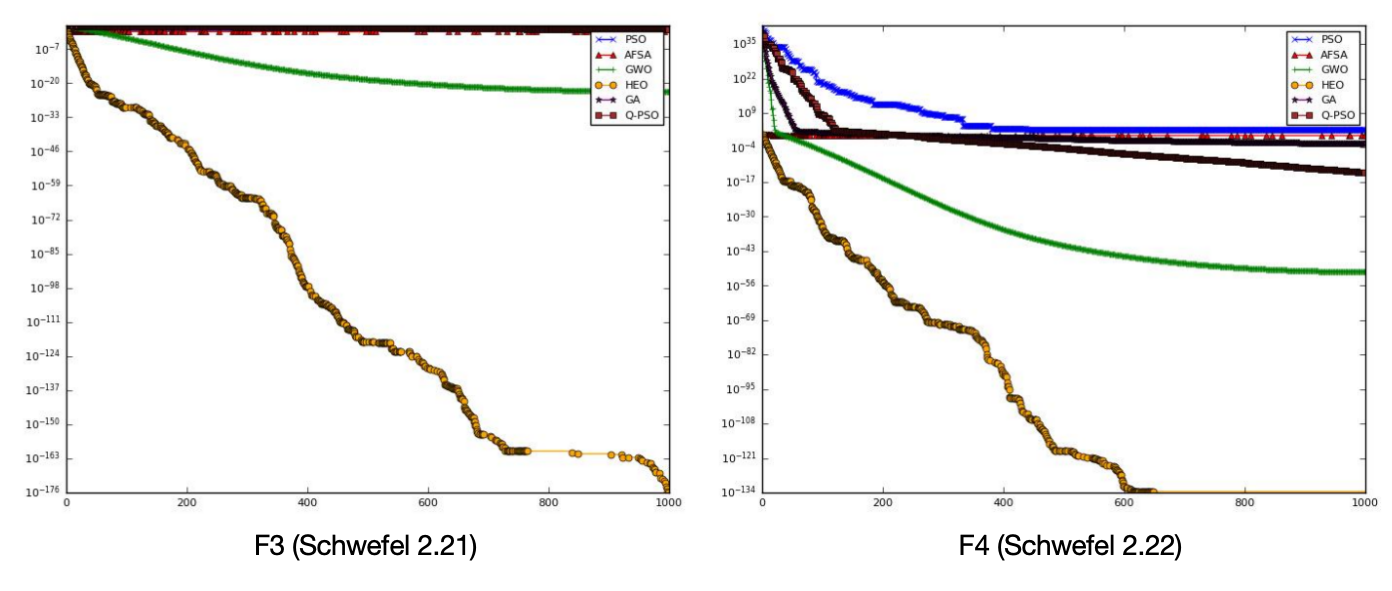}
    \caption{The cost variation in $F_{3}$ and $F_{4}$}
    \label{fig:F3-F4}
\end{figure}
For $F_{3}$, the Fig.\ref{fig:F3-F4} shows the fast convergence of HEO. Other algorithms including GWO seem only to find local optima of the function. Even the cost of GWO is 6.178e-24 (seen in Table \ref{tab:4.3}), which cannot be seen as an ideal situation.

For $F_{4}$, Fig.\ref{fig:F3-F4} also shows that the performances of HEO are much superior to GA, AFSA, and PSO. The HEO only gets around 1e-134 cost, and GWO achieves about 1e-50. The QPSO, AFSA, GWO, and GA express a similar trait on the plot, those algorithms all converge smoothly but stuck in optimums. Adversely, the oscillations of PSO could be caused by the complex landscape of the $F_{4}$ function, and the fluctuations of the HEO plot could be caused by its unique random mechanism, such as random skip and its escape behavior.

\begin{figure}[H]
    \centering
    \includegraphics[width=1\linewidth]
    {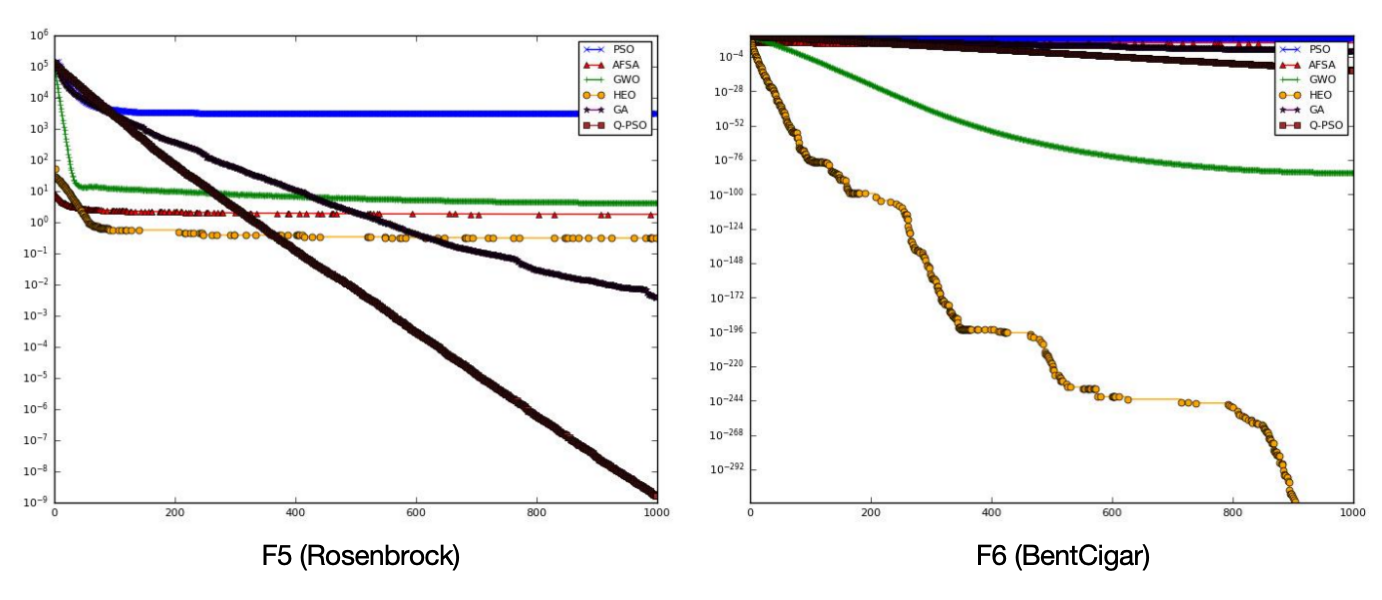}
    \caption{The cost variation in $F_{5}$ and $F_{6}$}
    \label{fig:F5-F6}
\end{figure}

For $F_{5}$ \cite{Rosenbrock1960}, Fig.\ref{fig:F5-F6} shows the majority of algorithms get trapped by the special curve structure, and only GA as well as QPSO gets a good convergence. Besides, the HEO also shows a reasonable cost that is lower than the others.

For $F_{6}$, Fig.\ref{fig:F5-F6} also shows the HEO finds the global optima, and the GWO algorithm also finds a good solution, as in Table \ref{tab:4.3}, we can observe a cost of 1.893e-85 cost. Other algorithms except GA do not have apparent tendency and they all converge too early as the plot shows.

\begin{figure}[H]
    \centering
    \includegraphics[width=1\linewidth]
    {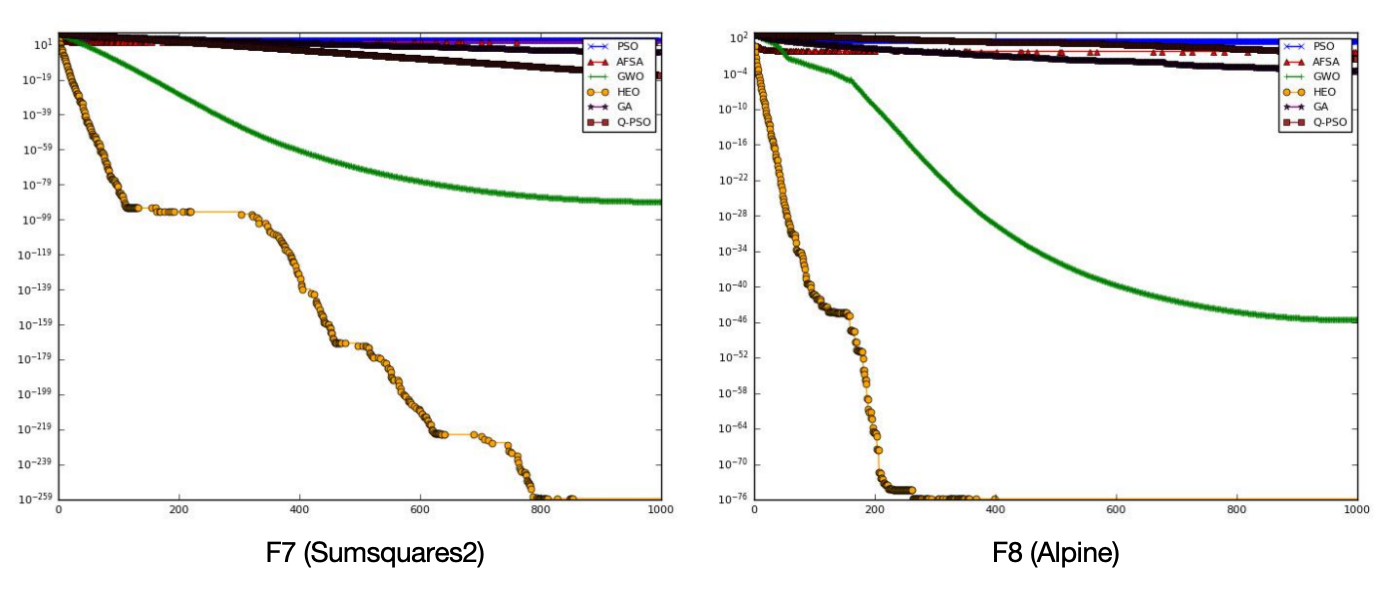}
    \caption{The cost variation in $F_{7}$ and $F_{8}$}
    \label{fig:F7-F8}
\end{figure}

For $F_{7}$, the Fig.\ref{fig:F7-F8} shows the HEO gets a good solution, and GWO follows after it. As for the convergence, HEO converges with the fastest rate, and GWO gets the second fastest rate, notwithstanding the AFSA getting the smallest cost in the first few iterations.

For $F_{8}$, Fig.\ref{fig:F7-F8} also shows the HEO gets the best performance compared to other algorithms, and GWO gets the second-best performance at around 1e-45 cost. Unlike HEO performs in other functions, the HEO gets a gentle plot shape, which might caused by the smooth landscape of the $F_{8}$.

\begin{figure}[H]
    \centering
    \includegraphics[width=1\linewidth]{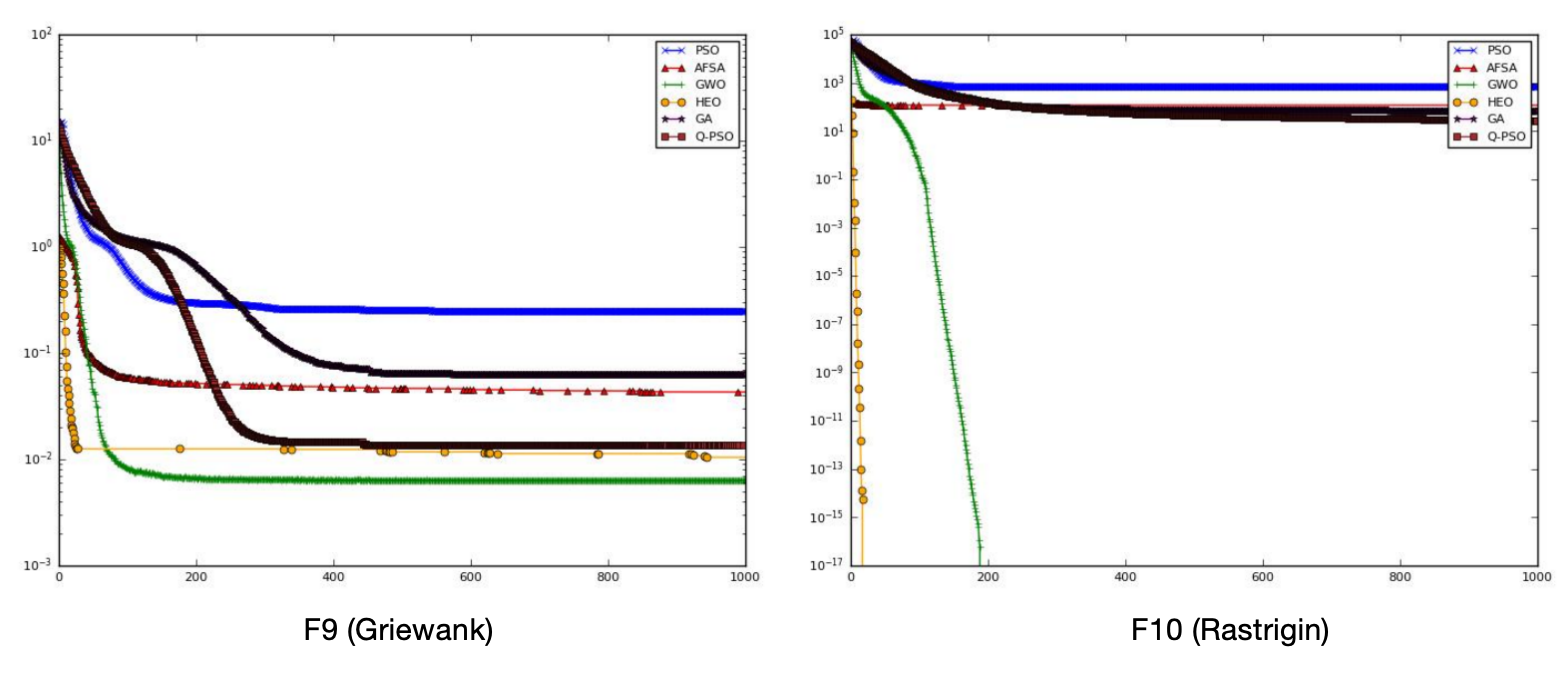}
    \caption{The cost variation in $F_{9}$ and $F_{10}$}
    \label{fig:F9-F10}
\end{figure}

For $F_{9}$ \cite{Griewank1981}, Fig.\ref{fig:F9-F10} illustrates all algorithms that encounter the difficulty of finding The global optima. The $F_{9}$, Griewank, is a complex function. With p=30, the algorithms are easy to trap in a single convex hull that does not contain the best solution. The HEO and GWO get approximate costs, but HEO converge faster and get slightly lower costs.

For $F_{10}$, Fig.\ref{fig:F9-F10} also shows that GWO as well as HEO both find the global optimum. For analyzing the convergence, the HEO finds the solution faster than the GWO.

\begin{figure}[H]
    \centering
    \includegraphics[width=1\linewidth]{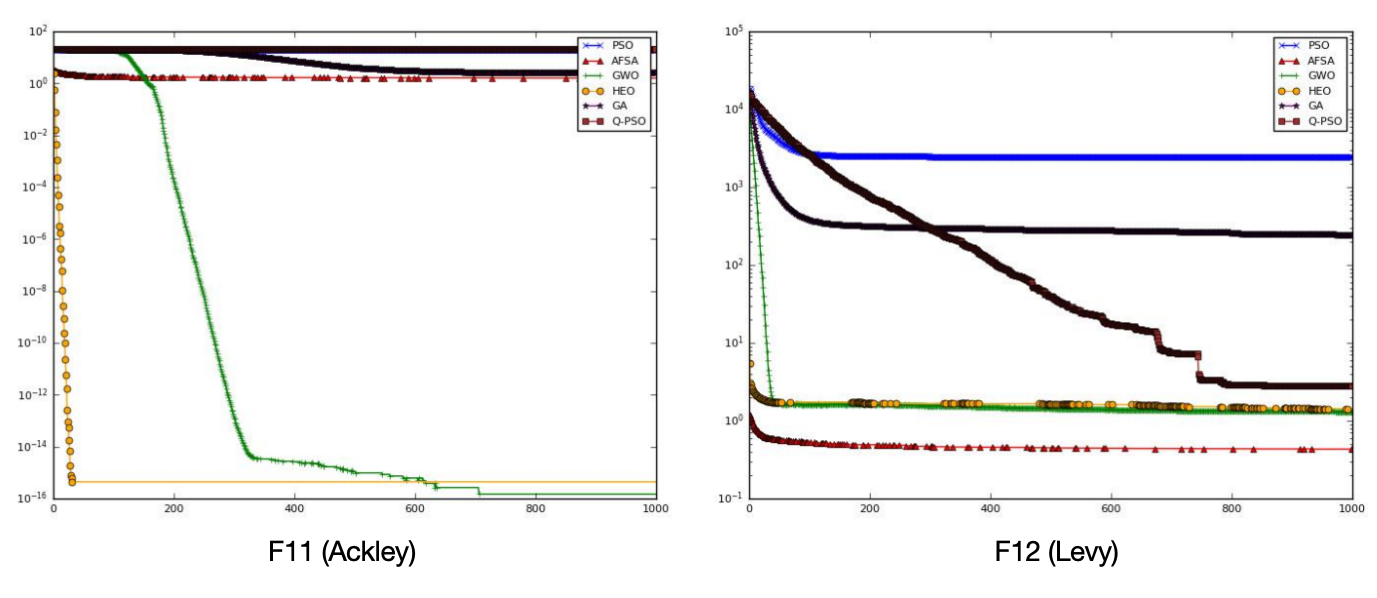}
    \caption{The cost variation in $F_{11}$ and $F_{12}$}
    \label{fig:F11-F12}
\end{figure}

For $F_{11}$, Fig.\ref{fig:F11-F12} shows that HEO and GWO both obtain a reasonable cost around 1e-16, but do not get full convergence. The possible reason for explaining this result could be the hollows of the Ackley function $F_{11}$ in high dimension\cite{cai2020solution}. Those hollow landscapes with the same cost might be the issue that prevents further convergence of algorithms.

For $F_{12}$, Fig.\ref{fig:F11-F12} also shows that the AFSA gets the lowest cost and then follows with GWO and HEO. All testing algorithms, including AFSA, do not perform well. The L$\acute{e}$vy functions is a folding bent shape function with some sheer landscapes, which might indicate a drawback for HEO.

\begin{figure}[H]
    \centering
    \includegraphics[width=1\linewidth]{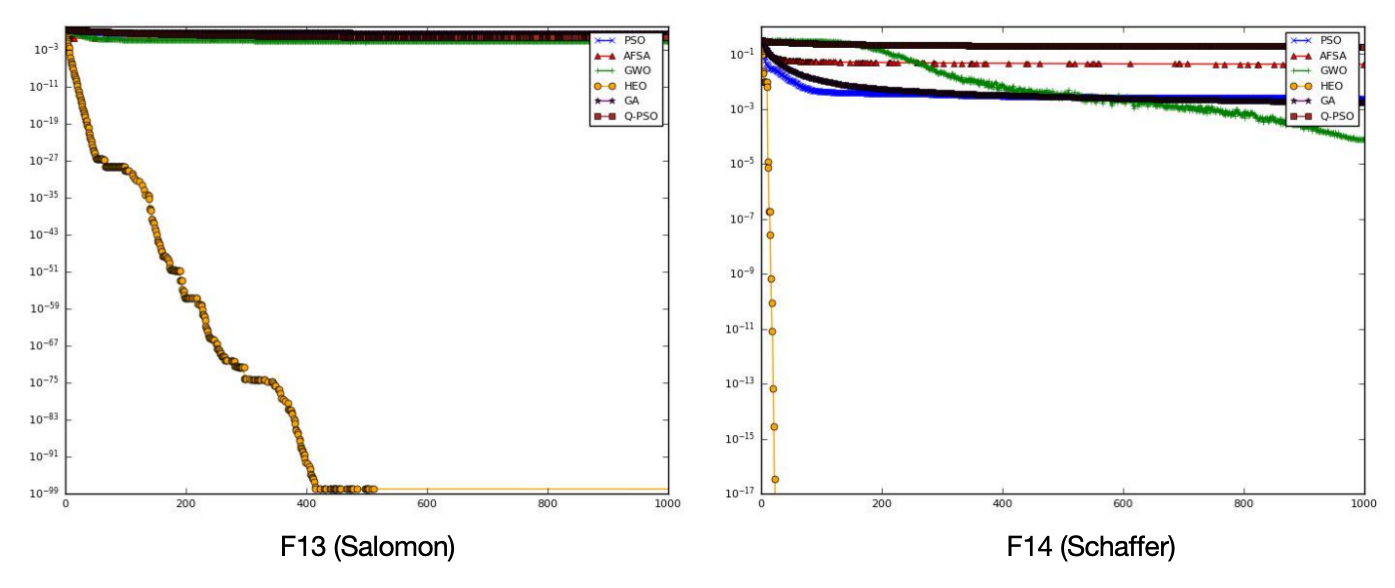}
    \caption{The cost variation in $F_{13}$ and $F_{14}$}
    \label{fig:F13-F14}
\end{figure}

For $F_{13}$, Fig.\ref{fig:F13-F14} shows the HEO gets the best result, which is near the global optimum. Salomon function $F_{13}$ is another sheer function but with an extent of convexity, making HEO able to find the solution.

For $F_{14}$, Fig.\ref{fig:F13-F14} also shows the HEO found the best solution in the first few iterations. In addition, the GWO has a small inclination to converge further and possibly to find the global optima.

In summary, from analysis of 14 benchmark functions, illustrates Halfway Escape Optimization (HEO) gets successful convergences on each case, which validates with the effectiveness of HEO. 

\subsection{Experimental Comparison}
\begin{table}[h]
\caption{Mean Costs of the Algorithms}%
\begin{tabular}{@{}llllllll@{}}
\toprule
&PSO&AFSA&GWO&\textbf{HEO}&GA&QPSO\\
\midrule
$F_{1}$&3.545e+02  &7.847e-01  &1.651e-91  &\textbf{0.000e+00}  &8.814e-07 &1.465e-19\\
         $F_{2}$&7.943e+00  &7.992e-01  &1.416e-01  &1.344e-03	  &1.742e-07 &\textbf{1.880e-19}\\
         $F_{3}$&1.420e+01  &4.495e-01  &6.178e-24  &\textbf{1.302e-176}	  &5.579e+00 &2.303e+00\\
         $F_{4}$&8.246e+02  &3.912e+00  &1.622e-51  &\textbf{2.498e-134}	  &4.592e-03 &3.713e-14\\
         $F_{5}$&3.168e+03  &2.235e+00  &4.271e+00  &3.112e-01	  &3.817e-03 &\textbf{1.603e-09}\\
         $F_{6}$&1.022e+09  &8.106e+05  &1.893e-85  &\textbf{0.000e+00}	  &1.077e+00 &6.314e-14\\
         $F_{7}$&2.470e+03  &3.790e+02	  &2.168e-89  &\textbf{3.531e-259}	  &4.843e-04 &4.084e-17\\
         $F_{8}$&3.381e+01  &7.514e-01	  &2.441e-46  &\textbf{1.299e-76}	  &3.697e-04 &4.062e-02\\
         $F_{9}$&0.247727  &0.053917	  &\textbf{0.006319}  &0.010466	  &0.064001 &0.013684\\
         $F_{10}$&715.590208  &137.192018  &\textbf{0.000000}  &\textbf{0.000000}	  &67.088411 &27.315077\\
         $F_{11}$&2.000e+01  &1.811e+00  &\textbf{1.480e-16}  &4.440e-16	  &2.665e+00 &2.088e+01\\
         $F_{12}$&2439.240184  &\textbf{0.544817}  &1.314077  &1.421046	  &245.547882 &2.834483\\
         $F_{13}$&4.046e+00  &2.954e-01  &5.659e-02  &\textbf{9.341e-99}	  &3.357e+00 &3.932e-01\\
         $F_{14}$&0.002472 &0.050465  &0.000076  &\textbf{0.000000}  &0.001709 &0.186564\\
\botrule
\end{tabular}
\label{tab:4.3}
\end{table}

The bold text in Table \ref{tab:4.3} is the smallest cost for one objective function. As the overall result that Table \ref{tab:4.3} shows, the HEO gets a better comprehensive performance than PSO, AFSA, GWO, GA, and QPSO in major benchmark functions $F_{1}$ to $F_{14}$, and get the minimum costs over 9 functions. On another side, HEO gets a relatively bigger cost in  $F_{2}$, $F_{5}$, and $F_{12}$, which might infer the disadvantages of using escape and skip mechanisms in HEO.

\begin{table}[h]
\caption{The Average Ranks of the Costs}%
\begin{tabular}{@{}llllllll@{}}
\toprule
&PSO&AFSA&GWO&\textbf{HEO}&GA&QPSO\\
\midrule
        Unimodal Functions&5.7142&4.5714&2.7142&\textbf{1.5714}&3.7142&2.7142\\
        Multimodal Functions&5.4285&3.5714&\textbf{1.5714}&\textbf{1.5714}&4.0000&4.1428\\
        Total Rank$(F_{1}-F_{14})$&5.5714&4.0714&2.1428&\textbf{1.5714}&3.8571&3.4285\\
\botrule
\end{tabular}
\label{tab:rank}
\end{table}

For a more precise analysis, the article performs a straightforward rank aggregation procedure on the data presented in Table \ref{tab:4.3}. Algorithms with lower associated costs are accorded higher ranks within this aggregation. The resulting aggregated rank data is displayed in Table \ref{tab:rank}. In Table \ref{tab:rank}, Halfway Escape Optimization (HEO) achieves the highest rank in both unimodal functions (encompassing $F_{1}-F_{7}$) and multimodal functions (encompassing $F_{8}-F_{14}$). This suggests that HEO exhibits similar performance compared to the Grey Wolf Optimizer (GWO) specifically in handling multimodal Functions, it also demonstrates competitive performance relative to the other six algorithms tested in unimodal functions, even in scenarios involving relatively complex functions.

\begin{table}[h]
\caption{Search Time(s/1000 iters) of the Algorithms}%
\begin{tabular}{@{}llllllll@{}}
\toprule
&PSO&AFSA&GWO&HEO&GA&QPSO\\
\midrule
        $F_{1}$&2.732496&136.057096&74.678300&
        18.509185&9.189708&25.709296\\
        $F_{2}$&3.307250&150.451374&76.191960&
        20.107816&10.300027&26.042222\\
        $F_{3}$&0.522904&36.712690&34.066304&
        8.800284&4.023997&10.214057\\
        $F_{4}$&1.587224&86.004913&72.936023&
        15.730473&8.095256&24.084705\\
        $F_{5}$&6.240265&237.248756&79.063583&
        25.920876&13.020591&29.193232\\
        $F_{6}$&2.916587&139.673755&75.215040&
        18.904487&9.627195&25.440708\\
        $F_{7}$&15.078621&509.393016&87.199845&
        43.967146&21.811757&37.721174\\
        $F_{8}$&1.059602&63.582046&43.069062&
        12.031251&6.521126&13.315242\\
        $F_{9}$&10.000640&315.942743&53.234886&
        31.335585&15.717625&22.645855\\
        $F_{10}$&10.626591&309.697071&83.785108&
        35.644457&17.775824&33.821670\\
        $F_{11}$&12.349613&381.549480&85.610975&
        38.228737&19.336686&34.786341\\
        $F_{12}$&7.995789&266.661428&50.095542&
        27.491028&14.171460&20.122009\\
        $F_{13}$&3.320846&45.687899&77.897484&
        19.997064&10.503721&26.490371\\
        $F_{14}$&7.923606&273.046009&51.550709&
        26.627504&14.288681&20.465703\\
\botrule
\end{tabular}
\label{tab:4.4}
\end{table}

For the efficiency, based on the data from Table \ref{tab:4.4}, the search time of five algorithms in these functions always shows a relationship that: $T_{PSO} < T_{GA}< T_{HEO}\approx T_{QPSO} < T_{GWO}  \approx T_{AFSA}$.  In conclusion, the HEO demonstrates an acceptable search time as well as a good performance among test functions except for $F_{12}$, but further validation of its effectiveness is still needed. For example, for a more comprehensive view, further tests could focus on testing HEO in other objective functions and various dimensions instead of just 30. This broadened experiment could give a more precise evaluation of the HEO.

\section{Stimulated Engineering Problems}
\subsection{The Pressure Vessel Design Problem}
\begin{figure}[H]
    \centering
    \includegraphics[width=0.9\linewidth]{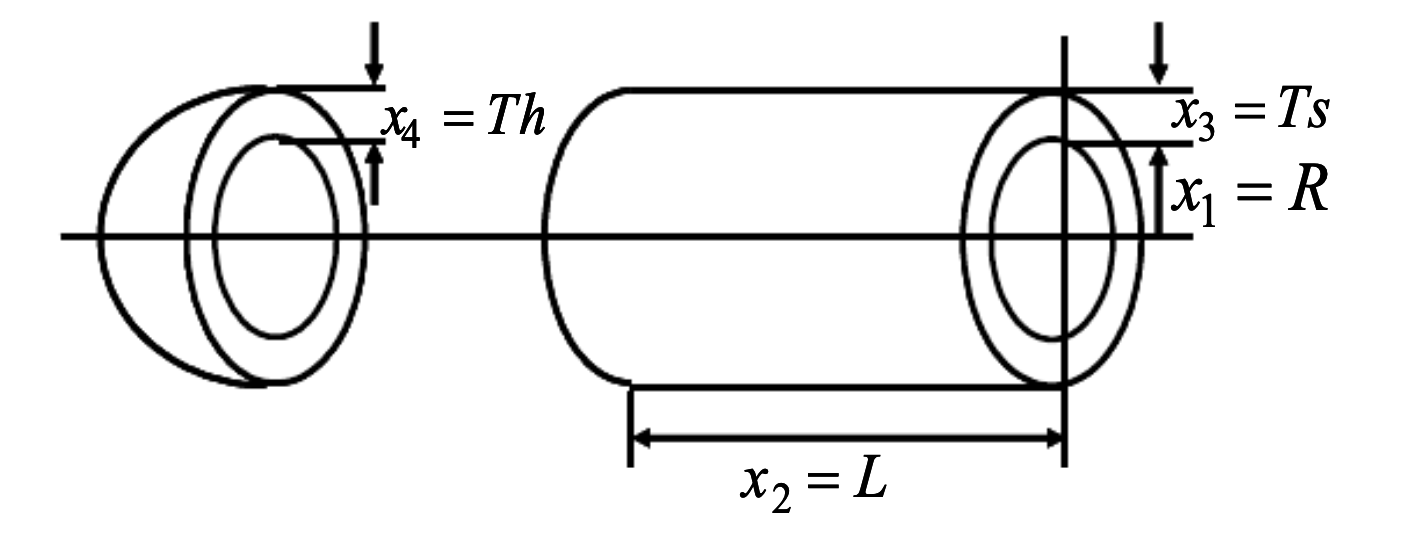}
    \caption{The Structure of the Vessel-\cite{vesselS}}
    \label{fig:vessel}
\end{figure}

\begin{figure}[H]
    \centering
    \includegraphics[width=0.9\linewidth]{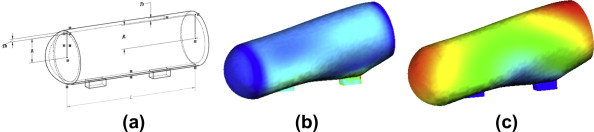}
    \caption{The Heatmaps of the Vessel-\cite{mirjaliliref}}
    \label{fig:heat}
\end{figure}

The pressure vessel design is a classical engineering problem, the objective is to change the structure of the vessel to reduce its manufacturing cost \cite{vessel}. The optimization algorithm tested in this problem could well validate its performance in complex constrained situations. As Fig.\ref{fig:vessel} and Fig.\ref{fig:heat} shown, the vessel composed by a cylinder shape of the structure welded with a sphere bottom. By changing from $x_1$ to $x_4$ values($x_1$:($R$-Radius of the Shell), $x_2$:($L$-Length of the Shell), $x_3$:($T_s$-Thickness of the Shell) ,$x_4$:($T$-Thickness of the dish end)), the task is to minimize the cost in equation\eqref{eq:26}. 

\begin{equation}
\label{eq:26}
\begin{aligned}
\\
&Consider:\\
&\bold{\Vec{x}}=[x_{1},x_{2},x_{3},x_{4}]\\
&Minimize:\\
&f(\bold{\Vec{x}})=0.6224x_{1}x_{2}x_{3}+1.7781x_{1}^{2}x_{3}+3.1161x_{2}x_{4}^2+19.84x_{4}x_{1}^2 \\
&Subject\quad to:\\
&g_1(\bold{ \Vec{x} })=-x_1+0.0193x_3\leq 0\\
&g_2(\bold{ \Vec{x} })=-x_3+0.000954x_3\leq 0\\
&g_3(\bold{ \Vec{x} })=-\pi x_3^{2}x_4+\frac{4}{3}\pi x_{3}^3+1296000\leq 0\\
&g_4(\bold{ \Vec{x} })=x_4-240\leq 0\\
&Variable\quad Range:\\
&0.00625\leq x_1 \leq 1.25\\
&0.00625\leq x_2 \leq 1.25\\
&40\leq x_3 \leq 200\\
&40\leq x_4 \leq 200\\
\end{aligned}
\tag{18}
\end{equation}

\begin{table}[h]
\caption{Pressure Vessel Design in 30 times Validation}%
\begin{tabular}{@{}llllllll@{}}
\toprule
&PSO&GWO&HEO&GA&\textbf{QPSO}\\
\midrule
        $\mu_{cost}$&9938.35&9197.10&8638.74&
        8757.88&\textbf{7680.54}\\
        $\sigma_{cost}$&1952.30&1890.72&1268.32&
        1533.92&\textbf{825.02}\\
        $T_{s}^*$&0.8125&1.25&1.25&
        1.0&0.9375\\
        $T_{h}^*$&0.4375&0.5625&0.5625&
        0.5&0.5\\
        $R^*$&40.3227&58.9299&57.6567&
        51.4577&88.9923\\
        $L^*$&200.0&40.4031&47.4311&
        48.3081&112.6784\\
\botrule
\end{tabular}
\label{tab:vessel}
\end{table}

\begin{figure}[H]
    \centering
    \includegraphics[width=1\linewidth]{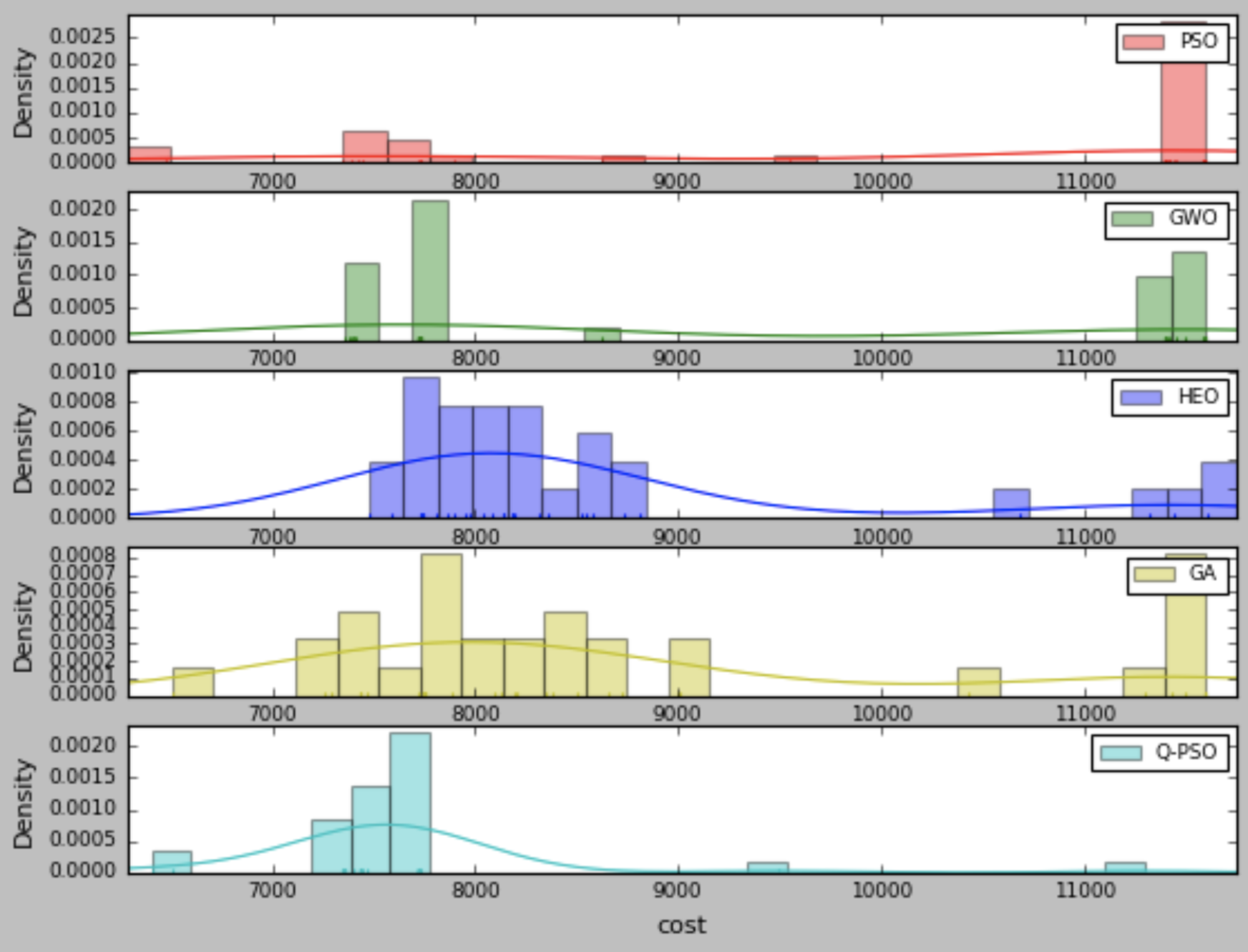}
    \caption{The Distribution of Cost for Pressure Vessel Design}
    \label{fig:dist1}
\end{figure}

As Table \ref{tab:vessel} and Fig.\ref{fig:dist1} illustrated, the HEO algorithm show a relativelyx small cost among the others, but larger than the cost of QPSO. The Pressure Vessel Design Problem is a linear combination problem with constraints, which is similar to the case of the Step Function. From the HEO's cost in the Step Function(or $F_2$) being larger than QPSO,it is not unexpected that HEO perform worse than QPSO in this problem, which might indicate HEO's weakness when solving the complex problem with simple constraints.

\subsection{Tubular Column Design Problem}
\begin{figure}[H]
    \centering
    \includegraphics[width=0.75\linewidth]{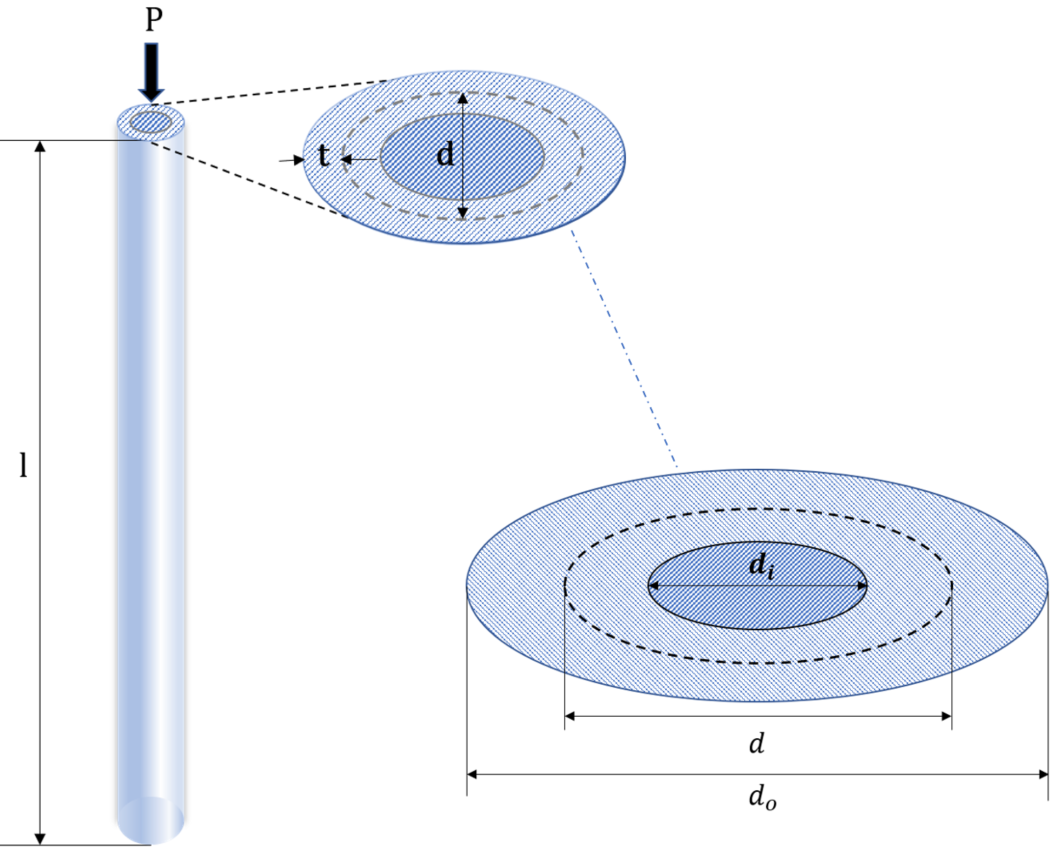}
    \caption{The Structure of the Tubular Column Design-\cite{TubularColumn}}
    \label{fig:col}
\end{figure}

As Fig.\ref{fig:col} shown, the Tubular Column this problem is structural part that is a cylindrical shaft, which may be solid, or made of a single circular cross-section and produced from metal, concrete, and other materials \cite{TubularColumn}. It is commonly used within constructing projects where it is required to support beams, as well as other elements, such as columns, and so on, and it is also used in bridges and other types of structures. 

Typically, tubular columns are superior and efficient compared to solid columns because they can resist torsion, bending, and shear force. The ideal is to minimize the cost of realising the column which is captured by the variables d, the average diameter of the column in centimeters and t the thickness of the column in centimeters.

In Table \ref{tab:col}, the paper sets 5 algorithms with the same parameter as the previous tests in the experiment(HEO with $c_{max}=1$), the AFSA does not show any inclination to converge to test engineering problems; hence, relevant data are eliminated in the table. The HEO has the smallest mean value of cost compared to other algorithms in 30 times validations, might illustrate its good performance in the tubular column problem,or relatively simple function with complex constraints. As for the deviations, the GA get the smallest one,indicating its stability to some extent.

\begin{equation}
\label{eq:27}
\begin{aligned}
\\
&Consider:\\
&\bold{ \Vec{x} }=[x_{1},x_{2}]\\
&Minimize:\\
&f(\bold{ \Vec{x} }) = 9.8x_{1}x_{2}+2x_{1}\\
&Subject\quad to:\\
&g_1(\bold{ \Vec{x} }) = \frac{P}{\pi x_{1}x_{2}\chi_{y}} - 1\leq 0\\
&g_2(\bold{ \Vec{x} }) = \frac{8PL^{2}}{\pi^{3}Ex_{1}x_{2}(x_{1}^{2}+x_{2}^{2})} - 1\leq 0\\
&g_3(\bold{ \Vec{x} }) = \frac{2}{x_{1}} - 1\leq 0\\
&g_4(\bold{ \Vec{x} }) = \frac{x_{1}}{14} - 1\leq 0\\
&g_5(\bold{ \Vec{x} }) = \frac{0.2}{x_{2}} - 1\leq 0\\
&g_6(\bold{ \Vec{x} }) = \frac{x_{2}}{8} - 1\leq 0\\
&Variable\quad Range:\\
&2\leq x_{1}\leq 14\\
&0.2\leq x_{2}\leq 0.8\\
\end{aligned}
\tag{19}
\end{equation}

Among them in equation\eqref{eq:27}, $P=2300 $ means compressive load with unit $kg_f$, $chi_{y}=450 $(is yield stress, has unit $kg_f/cm ^ 2 $), $E=0.65*10 ^ {6} $ is elasticity,unit is $kg_f/cm ^ 2 $),as for $L=300 $, is length of column with unit $cm $.

\begin{table}
\caption{Tubular Column Design in 30 times Validation}%
\begin{tabular}{@{}llllllll@{}}
\toprule
&PSO&GWO&\textbf{HEO}&GA&QPSO\\
\midrule
        $\mu_{cost}$&32.105096&32.147502&\textbf{31.704160}&
        32.224944&32.144308\\
        $\sigma_{cost}$&0.419599&0.366717&0.760805&
        \textbf{0.015721}&0.369621\\
        $d^*$&7.1031&7.1013&7.1477&
        8.1346&7.1013\\
        $t^*$&0.229&0.2295&0.2279&
        0.2&0.2292\\
\botrule
\end{tabular}
\label{tab:col}
\end{table}

\section{Model-based Optimization for Real-time Problem}

The experimental results indicate that HEO outperforms other algorithms in terms of convergence speed and solution quality across a range of benchmark functions. The algorithm's ability to balance exploration and exploitation, coupled with its adaptive strategies, has resulted in competitive performance in challenging optimization landscapes. 
The discussion of stimulated engineering problems like Tubular Column Design Problem highlights the effectiveness of the HEO algorithm in general applications.

However, whether the performance of the HEO is enough to deal with real-time situations still questionable, the applications of HEO may have to be more focused on realms like industrial parts manufacturing instead of a more dynamic environment. In addition, some special functions like L$\acute{e}$vy or Step would still be an issue for HEO to explore solutions, the Pressure Vessel Design Problem also infer its shortage. 

For further testing of the HEO, the paper would use the tested optimization algorithms and Grid Search(because it is a good baseline for hyperparameter tuning\cite{Grid}) with 50 population size and 50 iterations(2500 size sample space for Grid Search) to optimize the Logistic Regression parameters for better performance in this classification data set, due to the Logistic Regression do not have any randomness(The definition of the Logistic Regression had shown in equation \eqref{eq:28})\cite{Logistic}. The first parameter is the Cost Parameter C in Logistic Regression, and the second one is the maximum iteration $i_{max}$ to minimize the mean square error shown in equation \eqref{eq:29} of the model. This method that find candidate solutions for model's parameter directly called the Model-based Optimization\cite{model-base}, Model-based Optimization provides a controlled and repeatable framework for testing, ensuring consistent and reliable results.

\begin{align*}
\label{eq:28}
h_{\theta}(\bold{x}) &= \frac{1}{1 + \exp{(-\bold{\theta}^T \bold{x})}} \\
J(\bold{\theta}) &= -\frac{1}{m} \sum_{i=1}^{m} [\bold{y}^{(i)} \ln{(h_{\bold{\theta}}(\bold{x}^{(i)}))} + (1 - \bold{y}^{(i)}) \ln{(1 - h_{\bold{\theta}}(\bold{x}^{(i)}))}] \\
\frac{\partial J(\bold{\theta})}{\partial \bold{\theta}_j} &= \frac{1}{m} \sum_{i=1}^{m} (h_{\bold{\theta}}(\bold{x}^{(i)}) - \bold{y}^{(i)}) \bold{x}_j^{(i)}
\tag{20}
\end{align*}

\begin{equation}
\label{eq:29}
\begin{aligned}
\\
&Consider:\\
&\bold{ \Vec{x} }=[x_{1},x_{2}]\\
&Minimize:\\
&f(\bold{ \Vec{x} }) =\sum \frac{((\beta_1 x_1 + \beta_2 x_2)-y)^2}{2}\\
&Variable\quad Range:\\
&1e-16 \leq x_{1}\leq 100\\
&1 \leq x_{2}\leq 100\\
\end{aligned}
\tag{21}
\end{equation}

Rice industry is world-wide and have a demand to selecting out the ones that satisfy the various quality criteria of countries\cite{rice-world}. The paper by Kay CINAR and Murat KOKLU suggest a new tabular dataset extracted from image characteristics to identify two rice species Osmancik-97 and Cammeo \cite{rice},a binary classification contains 2180 samples of Osmancik and 1630 for Cammeo. The performance of the models shown in the Table \ref{tab:rice}, with the same ratio(test set:train set is 0.25:0.75), the best model with HEO has a performance(accuracy: 0.9328, f1-score: 0.9328) that exceeds the record (accuracy: 0.9302, f1-score: 0.9180) in the original paper.The Evaluation Metrics including:

\begin{equation}
\begin{aligned}
\text{Accuracy} & = \frac{TP + TN}{TP + TN + FP + FN} \\
\text{Sensitivity} & = \frac{TP}{TP + FN} \\
\text{Specificity} & = \frac{TN}{TN + FP} \\
\text{Precision} & = \frac{TP}{TP + FP} \\
\text{Recall} & = \frac{TP}{TP + FN} \\
\text{F1-score} & = 2 \cdot \frac{Precision \cdot Recall}{Precision + Recall} \\
\end{aligned}
\tag{22}
\end{equation}

Where:
\\
- $TP$ represents the number of true positives.
\\
- $TN$ represents the number of true negatives.
\\
- $FP$ represents the number of false positives.
\\
- $FN$ represents the number of false negatives.

\begin{figure}[H]
    \centering
    \includegraphics[width=0.7\linewidth]{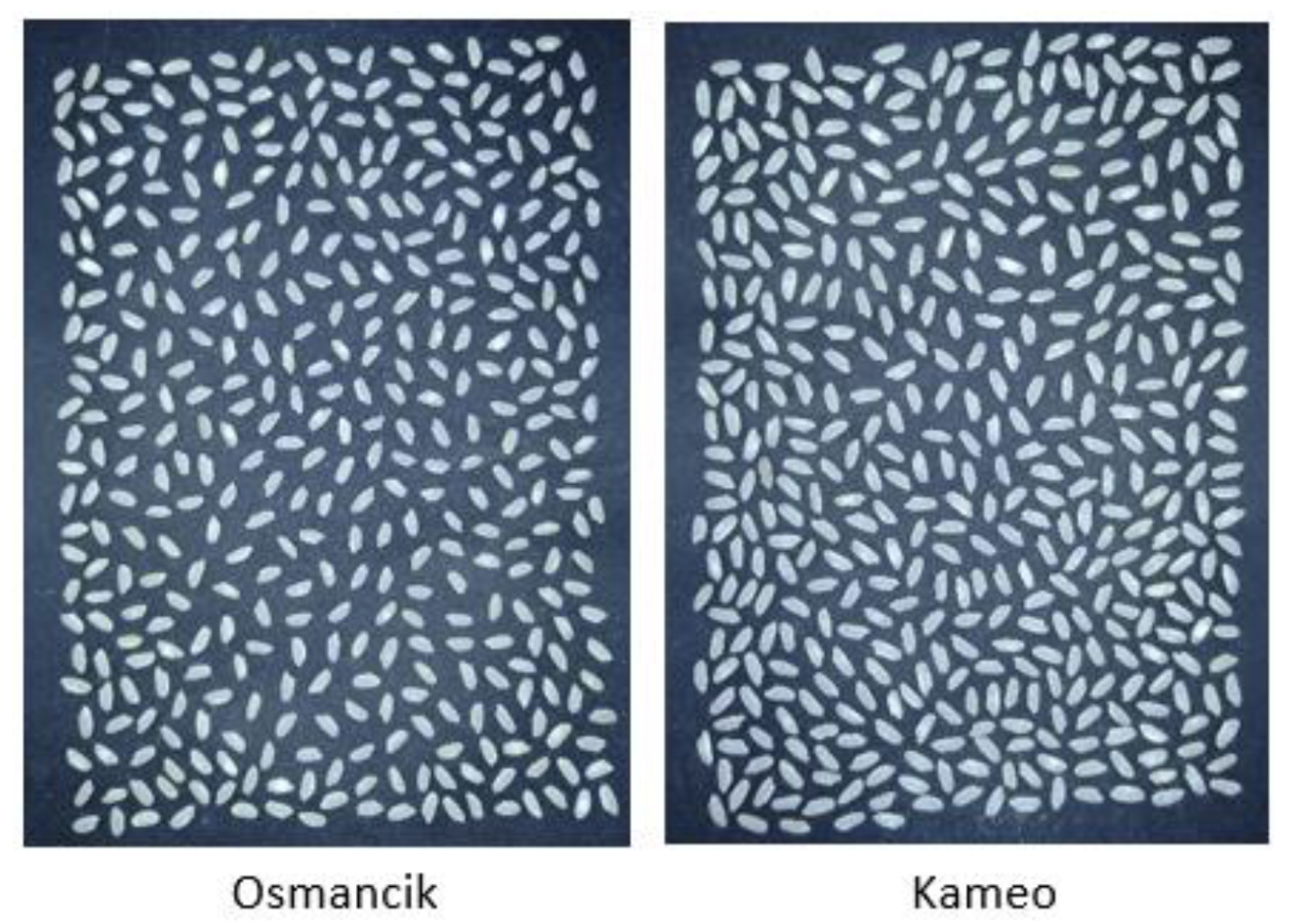}
    \caption{The raw sample of the Rice Dataset-\cite{rice}}
    \label{fig:rice}
\end{figure}

\begin{table}[h]
\caption{Rice Classification in Test Set}%
\begin{tabular}{@{}lllllll@{}}
\toprule
&PSO&GWO&\textbf{HEO}&GA&QPSO&GridSearch\\
\midrule
        Accuracy&0.9317&0.9296&\textbf{0.9328}&0.9317&0.9296&0.9296\\        Sensitivity&\textbf{0.9418}&0.9381&\textbf{0.9418}&\textbf{0.9418}&0.9381&0.9381\\
        Specificity&0.9181&0.9181&\textbf{0.9205}&0.9181&0.9181&0.9181\\
        Precision&0.9317&0.9296&\textbf{0.9328}&0.9317&0.9296&0.9296\\ 
        Recall&0.9317&0.9296&\textbf{0.9328}&0.9317&0.9296&0.9296\\ 
        F1-score&0.9317&0.9296&\textbf{0.9328}&0.9317&0.9296&0.9296\\ 
        SearchTime(s/it)&119.76&138.67&79.32&140.98&120.29&\textbf{22.28}\\
        $x_{1}(C)$&15.8281&11.6165&7.2574&7.6160&15.2641&6.0000\\
        $x_{2}(i_{max})$&60&74&77&62&74&51\\
\botrule
\end{tabular}
\label{tab:rice}
\end{table}

\section{Conclusion}
\label{cpt:con}
Through its quantum-inspired metaheuristic, Halfway Escape Optimization (HEO) demonstrates adaptability, efficiency, and effectiveness in tested problems. The algorithm's unique energy-driven behavior, vibration strategies, and exploratory mechanisms contribute to its ability to balance exploration and exploitation, leading to fast convergence rates and high-quality solutions.

Moreover, the adaptability of the HEO algorithm allows it to be applied to dynamic environments where the objective functions or constraints may change over time. By continuously adapting its search strategy and updating the position of particles, HEO can effectively track and respond to changing optimization landscapes in different engineering problems, ensuring that high quality solutions are maintained even in dynamic scenarios.

Overall, the HEO has demonstrated its stable effectiveness in 14 benchmarks functions and shown adaptability in addressing general optimization like the Pressure Vessel Design Problem, the Tubular Column Design Problem and the Model-based Optimization Problem. Future work could focus on further refining the algorithm's parameters and exploring its application in other practical optimization problems like path planning\cite{path}, circuit planning\cite{circuit}. The algorithm could also be extended to address multi-objective optimization problems and dynamic environments or change the strategy of escape and skip, further enhancing its versatility and applicability.

\section*{Declaration of competing interest}
The authors declare that they have no known competing financial interests or personal relationships that could have appeared to influence the work reported in this paper.

\section*{CRediT authorship contribution statement}
\textbf{Jiawen Li:}
Conceptualization, Methodology, Writing-Original Draft, Writing-Editing.
\textbf{Anwar PP Abdul Majeed:} Supervision, Writing-Review, Validation
\textbf{Pascal Lefevre:} Supervision, Writing-Review, Validation
\\
 
\section*{Data availability}
\sloppy
The experiment is data-available for replications, provide the source code of HEO in: \url{https://github.com/Spike8086/HEO.git}

\bibliography{sn-bibliography}

\end{document}